\documentclass{article}

\usepackage[preprint]{neurips_2019}

\usepackage[utf8]{inputenc} 
\usepackage[T1]{fontenc}    
\usepackage[draft]{hyperref}       
\usepackage{url}            
\usepackage{booktabs}       
\usepackage{amsfonts}       
\usepackage{nicefrac}       
\usepackage{microtype}      

\usepackage{graphicx}
\usepackage{xr}

\title{Continual Learning Using World Models for Pseudo-Rehearsal}

\author{
Nicholas Ketz\\
Information and Systems Sciences Laboratory\\
Center for Human Machine Collaboration \\
HRL Laboratories \\
Malibu, California 90265\\
\texttt{naketz@hrl.com} \\
\And
Soheil Kolouri \\
Information and Systems Sciences Laboratory\\
Center for Human Machine Collaboration \\
HRL Laboratories \\
Malibu, California 90265\\
\texttt{skolouri@hrl.com} \\
\And
Praveen Pilly\\
Information and Systems Sciences Laboratory\\
Center for Human Machine Collaboration \\
HRL Laboratories \\
Malibu, California 90265\\
\texttt{pkpilly@hrl.com}
}

\begin{document}

\maketitle

\begin{abstract}
The utility of learning a dynamics/world model of the environment in reinforcement learning has been shown in a many ways.  When using neural networks, however, these models suffer catastrophic forgetting when learned in a lifelong or continual fashion.  Current solutions to the continual learning problem require experience to be segmented and labeled as discrete tasks, however, in continuous experience it is generally unclear what a sufficient segmentation of tasks would be.  Here we propose a method to continually learn these internal world models through the interleaving of internally generated episodes of past experiences (i.e., pseudo-rehearsal).  We show this method can sequentially learn unsupervised temporal prediction, without task labels, in a disparate set of Atari games.  Empirically, this interleaving of the internally generated rollouts with the external environment's observations leads to a consistent reduction in temporal prediction loss compared to non-interleaved learning and is preserved over repeated random exposures to various tasks.  Similarly, using a network distillation approach, we show that modern policy gradient based reinforcement learning algorithms can use this internal model to continually learn to optimize reward based on the world model's representation of the environment.     
\end{abstract}

\begin{figure}[t!]
    \includegraphics[width=\columnwidth]{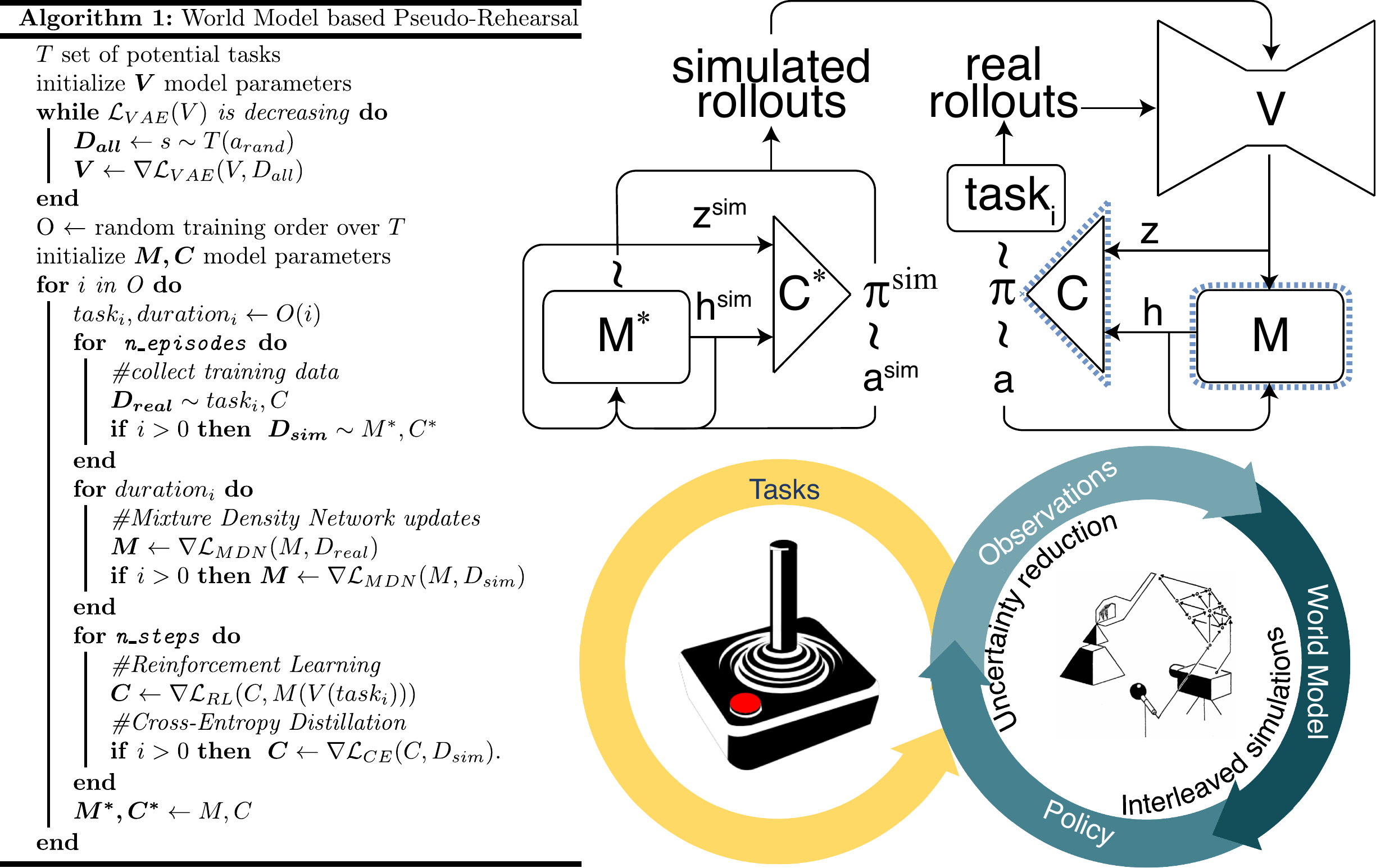}
    \caption{
Sequentially learning a set of tasks is done by iteratively training an internal world model $M$ and an agent $C$.  These neural networks are optimized based on samples from the current task interleaved with samples from simulated past experiences generated by preserved copies of those networks, $M^*$ and $C^*$.  
    }
     \small\textsuperscript{} Figure adapted from \citet{ha2018,kaiser2019,schmidhuber1990} 
    \label{fig:Model}
\end{figure}

\section{Introduction}

The power to simulate the dynamics of a given environment provides a learning agent the ability to not only re-experience past episodes, but also to generate potentially unseen experiences in preparation for encountering them.  This idea has its foundations in model-based reinforcement learning, however modern explorations have taken on many forms.  One proposed approach seeks to learn an unsupervised temporal prediction model that compresses the complete history of ($state_t, action_t, reward_t, state_{t+1}$) transitions an agent experiences \citep{schmidhuber2015,schmidhuber1990}.   This `world model' is then provided to the agent as a tool to better inform its decision making process in various ways.  Recent explorations of this theoretical framework have shown the approach is feasible at least within a single environment, and have illustrated the potential for training agents on simulated rollouts of possible episodes  from the learned world model in an entirely off-line fashion \citep{ha2018, kaiser2019}.

One critical aspect of this framework, which has yet to be fully addressed, is the need to continually learn in the potentially very different domains of the environment the agent is experiencing.  Particularly when using neural networks, the world model learned in this fashion would be subject to catastrophic forgetting when an incomplete history of all previous transitions are stored (which inevitably would be the case in a true lifelong learning scenario).  \citet{schmidhuber2015} suggests some ideas for how to address this, however they remain under-specified and untested to date. 

Modern approaches to the catastrophic forgetting problem propose to learn various forms of plasticity parameters, such that once a set of weights are learned for a given distribution of input samples (e.g., a particular task) those weights are made static as new distributions of input samples arrive \citep{kirkpatrick2017,zenke2017}.  Intrinsic to these approaches is the necessity to segment and label collections of input samples into discrete sets or `tasks'.  This is both difficult to do in the continuous flow of experience, and undesirably inflexible given the potential benefit of  transferring learned knowledge from one task to another.

Previous work in psychology and neuroscience have developed theories for how mammalian brains can perform this type of continual learning.  One of the more prominent computational theories is the idea of Complimentary Learning Systems \citep{mcclelland1995,oreilly2014}, which posits that recent experiences are `replayed' during both sleep and quiet rest.  This allows the brain, or learning system, to interleave previous experiences so they can be slowly integrated into a more comprehensive and robust internal representation.  This in-turn inspired connectionist models that used a form of this replay, referred to as `pseudo-rehearsal', to preserve previous learned weights  \citep{robins1995,ans2004}. The utility of such a mechanism to modern learning systems has been previously posited \citep{kumaran2016}, however the limit and scope of its applications have yet to be fully understood.

Here we explore a potential solution to preserve the unsupervised learning of a world model like architecture by adapting the strategy of pseudo-rehearsal \citep{robins1995,ans2004}.  Fundamentally this is accomplished by having the world model generate rollouts of previous experiences that are then interleaved with new experiences, thereby making the input distribution reflect both past and present experiences.  Critically, we show this method capable of preserving the previously learned experiences without the need for segmenting experience into discrete tasks.  To do this we use the original published version of the world model architecture from \citet{ha2018} trained over a random sequence of exposures to a set of Atari 2600 games in an unlabeled fashion.

The major contributions of this paper are summarized as:
\begin{itemize}
  \item Illustrating pseudo-rehearsal as a label-free approach to continual learning in pixel-based environments.
  \item Providing foundations for continual learning in model-based reinforcement learning with policy gradient methods.
  \item Exploring the content of pseudo-rehearsals and their relationship to performance retention and transfer.
\end{itemize}

\section{Related Work}

The original ideas surrounding pseudo-rehearsal were first developed within a relatively simple connectionist framework that relied on input and output to be sparse, low dimensional binary patterns.  In general, these methods used random input patterns to generate corresponding output patterns to create a set of input/output pairs that can be interleaved with new samples to preserve the current state of the network \citep{robins1995}.  Later work explored the idea of capturing a whole sequence of experiences with a single pseudo-sample \citep{ans2004}. These methods were both inspired by and inspiring to theories of the how biological systems continually learn, namely the Complimentary Learning Systems theory.  This theory posits that the hippocampus replays recent experiences so that the slower learning neo-cortex can consolidate that information into a more stable and robust representation \citep{oreilly2014}.

This idea of replay based preservation of past experiences has taken root in recent work looking to accomplish the goal of sequentially learning multiple datasets.  Most of this has been focused on classification tasks, and uses a form of pseudo-rehearsal referred to as `deep generative replay' \citep{shin2017}.  The basic approach is the same, however the input/output patterns can be non-sparse continuous values, including pixel-based representations.  The architecture used is a generative model of some type and is usually a Generative Adversarial Network \citep{goodfellow2014}.  Pushing this approach further, and taking inspiration from the Complimentary Learning Systems framework, a dual memory system used a form of deep generative replay for continual learning on a digits classification tasks \citep{kamra2017}. Notably, this implementation showed increased accuracy and backward transfer compared to Elastic Weight Consolidation, however the apparent difference in tasks tested was relatively small compared to the arbitrary set of Atari games used in the current work.  

The Compress and Progress framework \citep{schwarz2018} can continually train reinforcement learning agents in a series of complex pixel-based environments using a more scalable form of Elastic Weight Consolidation \citep{kirkpatrick2017} along with Policy Distillation \citep{hinton2015} to iteratively transfer learned policies into a single network.  As mentioned above these plasticity based preservation methods require task labels to avoid catastrophic forgetting.  Recent work, referred to as the RePR model (Reinforcement-Pseudo-Rehearsal), has shown that the pseudo-rehearsal approach is also viable in reinforcement learning \citep{atkinson2018}.  Using a GAN based generative architecture paired with a modified DQN this approach used pseudo-rehearsal to iteratively maintain a single generative network trained on self-generated pseudo-samples and samples from an expert generative network trained on a specific task.  For the integrated policy, similar to Progress and Compress \citep{schwarz2018}, a single agent network was learned by using Policy Distillation to generate target output action distributions from both itself and an agent network trained on a specific task.  This approach is very promising for iterative training in reinforcement learning, however, it segments experience into tasks to train expert generative agent networks, and can not provide temporally connected samples essential for n-step gradient based learning, e.g., policy gradient approaches. 

Within the world model framework there has been recent work that illustrate its effectiveness within reinforcement learning.  In particular, \citet{ha2018} showed that this framework could be used, within a limited set of environments, to train evolved neural networks to achieve state of the art performance even when trained entirely on simulated rollouts.  Similarly, \citet{kaiser2019} adapted the original world model architecture for the Atari environment and illustrated its sample efficiency relative to modern model-free reinforcement learning algorithms.  Finally, an approach that similarly learns a latent world model used to inform action selection was able to learn simultaneously in 6 separate environments provided there were interleaved with each other \citep{hafner2018}.

\section{Continual Learning of World Models}

Our main focus in this work is testing the hypothesis that interleaving pseudo-rehearsal based replays can be sufficient to preserve the learned knowledge within a world models framework. As illustrated in Figure \ref{fig:Model}, and similar to \citet{ha2018}, a Variational Auto Encoder ($V$) is trained to compress a high dimensional input (e.g. images) into a smaller latent space ($z$) while also allowing for a reconstruction of that latent space back into the high dimensional space. This latent space representation is then fed into a temporal prediction network ($M$) that is trained to predict one time step into the future. The output of these networks is then used as a latent state-space for a reinforcement learning based controller ($C$).  A illustration of the network architecture used in this work is shown in Figure \ref{fig:Model} and more detailed in Supplemental Figure \ref{fig:full_arch}.

\subsection{Model Architecture}

The $V$ network learns to both encode and reconstruct observed samples into a latent embedding by optimizing  a combination of reconstruction error of the samples from the embedding back into the original observation space, and the KL Divergence of the samples from the prior ( $ \mathcal{N}(\mu=0,\sigma=I)$) on the embedding space those samples are encoding into; this framework is generally known as a Variational Auto-Encoder (VAE) \citep{kingma2013}.  In this work we use a convolutional VAE with the same architecture as \citep{ha2018}, where the input image is passed through 4 convolutional layers (32, 64, 128, and 256 filters, respectively) each with a 4x4 weight kernel and a stride of 2.  Finally, the output of the convolution layers is passed through a fully connected linear layer onto a mean and standard deviation value for each of the dimensions of the latent space, which is used to then sample from that space.  For reconstruction a set of deconvolution layers mirroring the convolution layers takes this latent representation as input and produces an output in the same dimensions as the original input.  All activation functions are rectified linear except the last layer which uses  a sigmoid to constrain the activation between 0 and 1.  

The $M$ network, also based on \citet{ha2018}, takes the latent space observation and passes it through an LSTM layer \citep{hochreiter1997}.  The LSTM output is then concatenated with the current action as input to a Mixture Density Network \citep{bishop1994}, which passes the input through a linear layer onto an output representation that is the mean ($\mu$) and standard deviation ($\sigma$) used to determine a specific Normal distribution ($\mathcal{N}$), and a set of $G$ mixture parameters ($\Pi$) used to weight those separate distributions in each of $V$'s latent space dimensions.  Two additionally output units are present, one for predicted reward and the other for the predicted episode termination probability.  In total that is $3GL+2$ output units, where $L$ is the size of the latent space.  See Supplemental section \ref{sec:m_desc} for a more complete description of $M$ optimization.

Finally, in departure from \citet{ha2018}, a stochastic gradient descent based reinforcement learning controller, $C$, was used for action selection.  The $C$ network takes as input the current hidden state ($h$) of the $M$ network concatenated with the current latent vector ($z$).  Internally there is a single fully connected 512 unit hidden layer that splits off to the Actor and Critic heads.  For learning we choose the A2C algorithm\footnote{code for this implementation adapted from https://github.com/ShangtongZhang/DeepRL}, the synchronous adaption of the original A3C algorithm \citep{mnih2016}, as we wanted to illustrate the effectiveness of our approach in a policy gradient based method, however, in theory any architecture would work. 

\subsection{World Model Based Pseudo-Rehearsal}

The training of these networks is usually done in sequence ($V$ then $M$ then $C$) and is entirely unsupervised (i.e. no labeled data is required). This framework allows for an offline simulation of experiences as the $M$ network can predict one time step into the future, and can use it's own predictions as the seed for the subsequent predictions provided some action derived from $C$.  After sampling a specific prediction $z^{sim}$ from the distribution of potential outcomes prescribed by the $M$ network's output 
these samples feedback as input and are allowed to rollout for several time-steps.  These rollouts used in this work for pseudo-rehearsal, samples approximating previous experiences, and interleaved with the real samples from the environment.  

The basic training algorithm for continual learning in this framework is outlined in Figure \ref{fig:Model}.  Continual learning in the $M$ and $C$ networks is done by first learning/providing an auto-encoder, $V$, that can embed high-dimensional samples from all potential environments into a sufficiently low-dimensional space.  If the input dimensions are already sufficiently small this embedding may potentially be unnecessary, however it is as yet unclear what a sufficiently low-dimensional space would be.   

The first iteration of training starts by sampling (with replacement) some particular task and duration to train on.  Data collection is done in this task using a random action selection policy and rollouts of $[[z_t,a_t,r_t,d_t]_{Tmax}]_{N}$ are saved.  Here for a given times-step $t$, $z_t$ is the latent representation of the current observation, $a_t$ is the chosen action, $r_t$ is the observed reward and $d_t$ is the binary done state of the episode.   For each task exposure, $N$ rollouts (1,000 in this work) are collected, where each rollout is allowed to proceed until $d_t$ is 1 or it reaches the $Tmax$ (1,000 in this work) maximum number of recorded time-steps.  The $M$ network is then optimized to perform 1-time-step prediction on these rollouts.  Following this, the $C$ network is trained to produce an action distribution $\pi$ such that sampled actions maximize the expected reward on the same task that $M$ was just trained on to.  This $C$ network uses the latent embedding of the current  observation ($z_t$) and the current hidden state ($h_t$) of the now trained $M$ network as input.  The $C$ network is consistently trained for \texttt{n\_steps} (1e6 in this work) within the current task.  This initial training is meant to generally mimic the Car Racing experiment of \citet{ha2018}.  At the end of each iteration of continual learning the current $M$ and $C$ networks are preserved as $M^*$ and $C^*$.  

At the start of the next iteration, a new task and duration are sampled, and the two networks are used to generate pseudo-samples to be interleaved with real samples from incoming new tasks.  First, a new set of real rollouts are generated using the current task processed through $M^*$ and $C^*$. Here, the $M^*$ network is provided a $V$ encoded observation from the current task which is in turn passed on to $C^*$ to produced a particular action, again yielding rollouts of the form: $[[z_t,a_t,r_t,d_t]_{Tmax}]_{N}$. Then, the simulated rollout generation process starts by picking a random point in the latent space here sampled based on the prior of the VAE (i.e., $\mathcal{N}(0,I))$, along with a zeroed out hidden state and a randomly sampled action.  Feeding this through $M^*$ produces the first simulated observation ($z^{sim}_0$ and hidden state $h^{sim}_0$ which are in turn provided to $C^*$ to yield the first distribution of potential actions $\pi_t^{sim}$ and the particular action sampled from that distribution $a_0^{sim}$.  Using the last sample as input to the $M^*$ network this process continues and the $[z^{sim}_t,a^{sim}_t,r^{sim}_t,d^{sim}_t, \pi^{sim}_t]$ tuples are stacked in time to produce simulated rollouts of pseudo-samples.  

These rollouts are in effect simulations of the tasks the network has already been exposed to, and can then be interleaved with new experiences to preserve the performance with respect to previous tasks in both the $M$ and $C$ networks.  Here, pseudo-rehearsal updates in $M$ are exactly the same as from real samples, just using the simulated rollouts in place of real rollouts.  Updates in the $C$ network are done using policy distillation, using a cross-entropy loss function with a temperature $\tau$ \citep{rusu2015}.  Specifically, provided a given simulated sample $z^{sim}_t$ as input, the temperature modulated softmax of $C$'s output distribution ($softmax(\pi_t/\tau)$) is forced to be similar to the temperature modulated softmax of the simulated output distribution ($softmax(\pi^{sim}_t/\tau)$)  from $C^*$.  Based on \citet{rusu2015} a constant $\tau$ of 0.01 was used in our experiments. 

\section{Experiments}

Testing of continual learning in the $M$ and $C$ networks was done by first generating 1000 rollouts from all potential tasks, which in this case were a set of 3 Atari games.  
Each random rollout was generated using a series of randomly sampled actions with a probability of 0.5 that a the last action will repeat.  These rollouts were constrained to have a minimum duration of 100 and maximum duration of 1000 samples.  
The first 900 of these rollouts, for each game, were used for training data and the last 100 are reserved for testing.  All image observations were reduced to 64x64x3 and rescaled from 0 to 1, and all games were limited to a 6 dimensional action space: ``NOOP", ``FIRE", ``UP", ``RIGHT", ``LEFT" and ``DOWN". Each game is run through the Arcade Learning Environment (ALE)  and interfaced through the OpenAI Gym \citep{bellemare13,brockman2016}.  All rewards were clipped as either -1, 0, or 1 based on the sign of the reward, the terminal states were labeled in reference to the ALE game-over signal, and a non-stochastic frame-skipping value of 4 was used.  The same environment parameters are used through-out the experiment. 

All training images are then fully interleaved to train a VAE ($V$ network) that can encode into and decode out of a 32 dimensional latent space. Training was done using a batch size of 32 and allowed to continue until 30 epochs of 100,000 samples showed no decrease in test loss greater than $10^{-4}$. 

\begin{figure}[ b!]
    \includegraphics[width=\columnwidth]{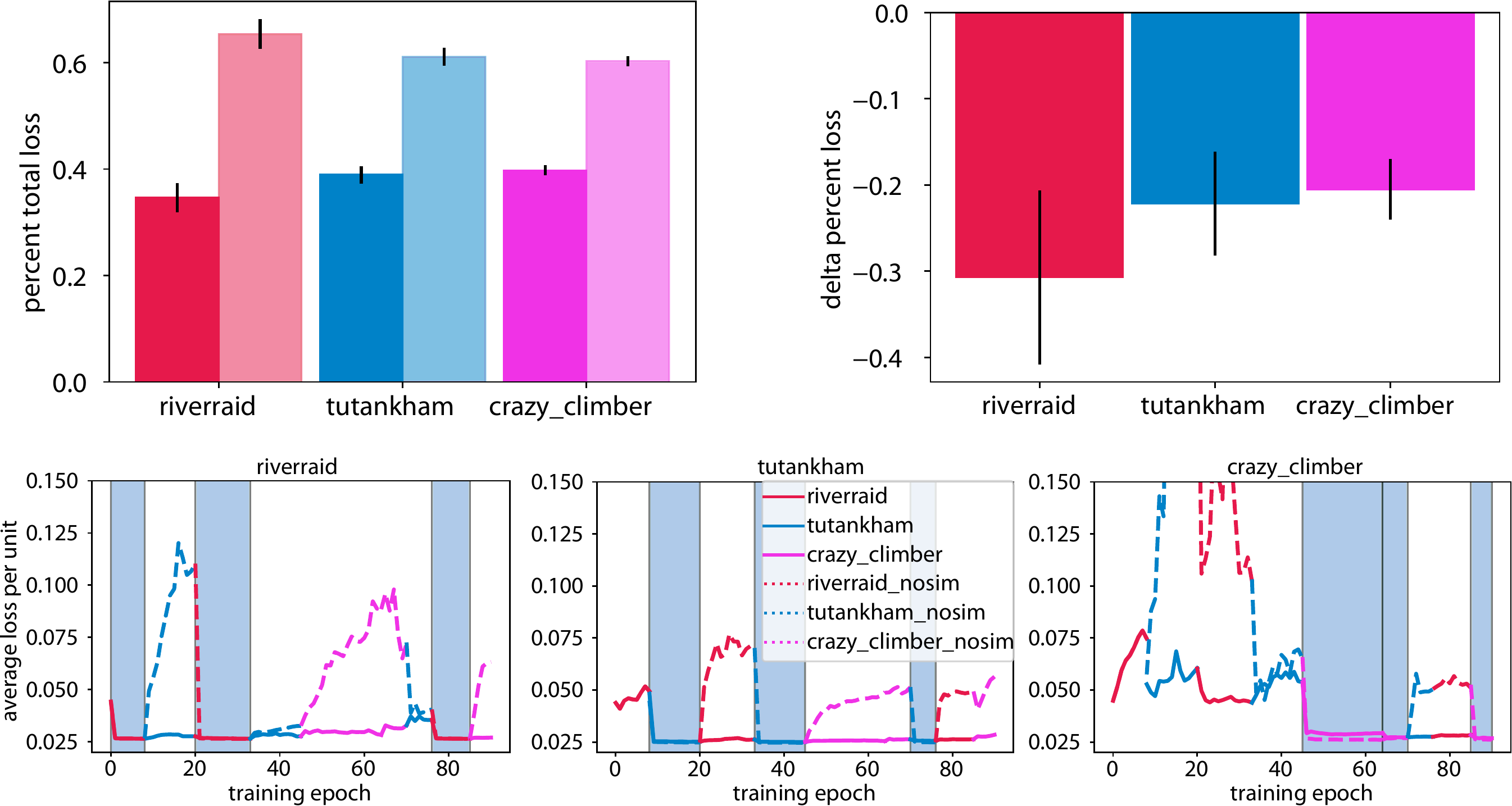}
    \caption{$M$ network loss over task exposures.  \textbf{Top-left} percent of total integrated loss over both experimental conditions (with and without pseudo-rehearsal), averaged over 10 random replications of task exposures.   Desaturated bars shows loss without pseudo-rehearsal.  Error bars are standard error of the mean. \textbf{Top-right}: pair-wise difference in total percent loss for each task.  Error bars are 95\% confidence intervals. \textbf{Bottom}: loss over training epochs for each task in a particular task exposure ordering.  Each plot corresponds to loss as measured in the hold-out set of the task label above it, line colors correspond to the task that is current being trained at a given epoch and overlaid boxes show when a given task is being exposed to its own training data. Solid lines show loss when using pseudo-rehearsal, dash lines without pseudo-rehearsal.     
    }
    \label{fig:cmb_loss}
\end{figure}

\begin{figure}[t!]
    \includegraphics[width=\textwidth]{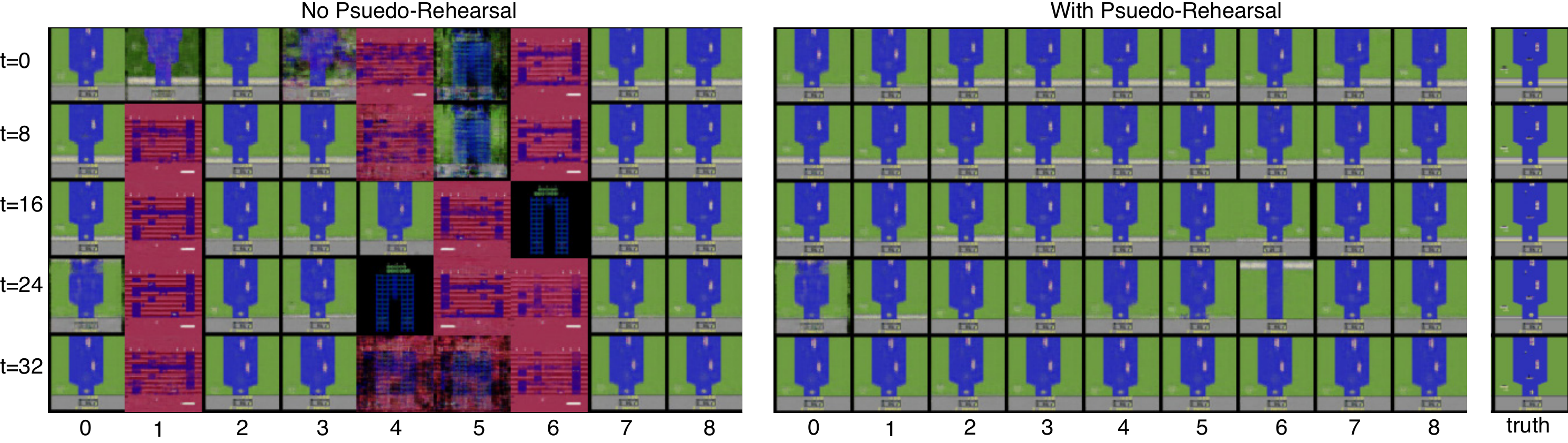}
    \caption{
    Reconstruction of test rollouts from riverraid using $M$ networks after training in each task exposure.  Left grid shows simulated rollouts when no pseudo-rehearsal was used in training, right grid shows simulated rollouts when trained with pseudo-rehearsal, and far right column shows real rollout from the environment.  Grid rows correspond to a given rollout's time steps,  columns  are specific rollouts generated after training is complete in each task exposure.
    }
    \label{fig:cmb_recon}
\end{figure}

Using this pre-trained $V$ network to encode the original rollouts into the latent space, the $M$ network is then trained over a series of randomly determined task exposures.  First, a random training order is determined such that all tasks have the same exposure to training, which is in total 30 epochs per task.  This total is split over the course of 3 randomly determined training intervals which each has a minimum of 3 epochs and a maximum determined by the floor of the total epochs left divided by the number of training exposures left for a given task.  The order over task exposures is then randomized with the exception that the first task and training duration (which has no pseudo-rehearsal) is always the same across random replications. Each epoch of training in the $M$ network is done using rollouts of length 32 in 100 batches of 16.  Once $M$ training is finished for a given task exposure, the output of this trained network is then used as input to the $C$ network for the same task.  In contrast to the random training duration of the $M$ network, training in the $C$ network was consistently set to 1 million frames per task exposure.

After every task exposure, the $M$ and $C$ networks are preserved as $M^*$ and $C^*$ as shown in Figure \ref{fig:Model}.  This pair of networks is then used to generate a set of 1000 simulated rollouts, or pseudo-samples.  For efficiency, these rollouts were saved into RAM at the start of each task exposure, but could easily be generated on-demand.  These generated rollouts are then interleaved with the next task's training set. Similarly, a set of 1000 real rollouts from the next task are generated using $M^*$ and $C^*$, mimicking the training scheme of \citet{kaiser2019}. 

Then on the next task exposure, the $M$ network is updated with 1 simulated rollout to 1 real rollout for the duration determined by the current task exposure.  Similarly, after training $M$, the $C$ network is allowed to explore the current task, however for every 30,000 frames from the current task, a batch of 30,000 simulated frames is trained using policy distillation.  $C$ training continues in each task exposure until 1e6 frames from the real task have been seen.  

The average loss per output unit in the $M$ network was used to asses performance, and median reward over 30,000 frames was used in the $C$ network.  A baseline measure of catastrophic forgetting was established by performing the same training as described above with no pseudo-samples interleaved.  Here, the area under the performance metric curve is integrated over all training epochs and divided by the sum over the two experimental conditions (training with and without pseudo-rehearsal) to achieve a percent performance that sums to one within each task. Performance statistics were calculated over 10 replications where a new random task exposure order was sampled for each replication.  Little to no hyper-parameter tuning was done in the $M$ or $C$ networks.  Here we were trying to replicate the $M$ and $V$ from \citet{ha2018} as close as possible, and were interested in $C$ as a form of performance metric rather than pursuing a SotA level of performance.

\section{Results}

\subsection{World Model}
Performance in the $M$ network (here the average loss per output unit) was assessed on the held out test-set of rollouts for each task and was done on all potential tasks at every epoch of training.  The bottom of Figure \ref{fig:cmb_loss} shows these performance curves for each of the 3 different Atari games as tasks. Solid lines show performance when simulated rollouts were interleaved during training, and dashed lines show performance when no interleaving of simulated rollouts occurred (with the label suffix of `\_nosim'). The different lines colors in each curve correspond to when the network was being training (as dictated in the legend) on a particular task.  Overlaid boxes indicate when a given task is engaged in training on its own data.  Clear catastrophic forgetting is seen in the non-interleaved case while relatively little reduction in loss is observed when simulated rollouts were interleaved.   The average percent loss, and pair-wise percent loss difference plots show that each task significantly more preserved when using pseudo-rehearsal.   

In Figure \ref{fig:cmb_recon}, reconstructions of test rollouts from the first task are shown across task exposures (see Supplemental Figure \ref{fig:sup_cmb_recon} for the other tasks).  This figure provides a heuristic for translating the change in loss observed in Figure \ref{fig:cmb_loss} into appreciable samples.  This also gives the reader an appreciation for the types of errors made when using pseudo-rehearsal vs. not, for a comparable level of loss.  Clear signs of catastrophic forgetting are seen in the reconstructed samples when simulated rollouts are not interleaved.  Conversely, the impact of the relatively small increased loss when using pseudo-rehearsal is seen in Figure \ref{fig:cmb_recon} to be relatively unintrusive when reconstructed into the original input space.  This increase does illustrate the potential for accumulating loss when scaling this method to 100s of tasks.   However, it can be seen in Figure \ref{fig:cmb_loss}) that each task showed significantly more accumulated loss when pseudo-rehearsal was not used. 

Further experiments using a larger set of tasks investigates transfer and continual learning in the $M$ network are explored in Supplemental section \ref{sec:sup_exp}.  These results suggest that this architecture can learn a disparate set of 10+ Atari games, however, some tasks start to show signs of reconstruction degradation if the duration between exposures is too long, or if their first exposure is not until late in training.  The complex dynamics of this relationship analyzed more completely in the Supplemental Material.

\begin{figure}[t!]
    \includegraphics[width=\textwidth]{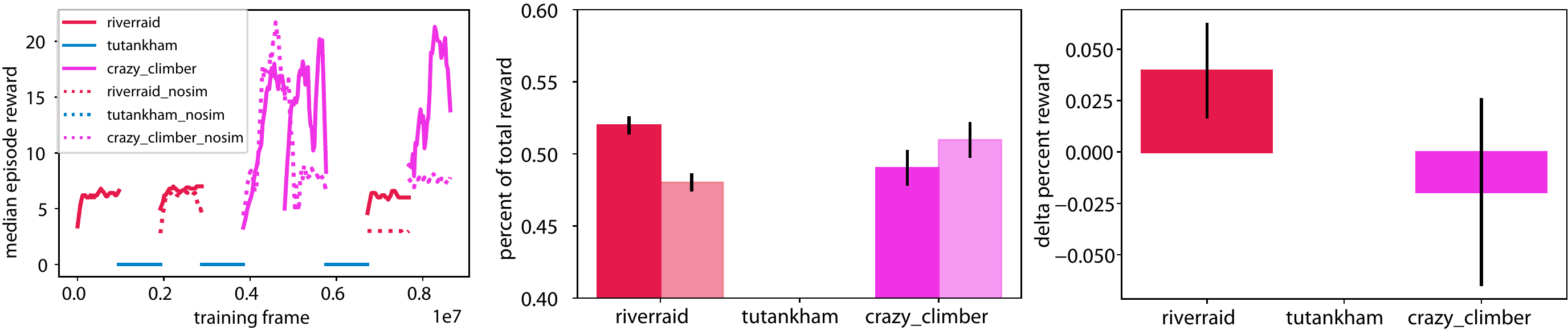}
    \caption{
    Cumulative reward in $C$ network over task exposures.  \textbf{Left}: median episode reward over 30,000 frames over task exposures for a particular replication, solid lines show percent reward when using pseudo-rehearsal, and dashed lines without (labeled with `\_nosim' suffix).  \textbf{Middle}: percent of total integrated reward averaged over 10 random replications, here desaturated bars show average percent reward when not using pseudo-rehearsal.  Error bars are standard error of the mean. plot shows pair-wise difference between percent reward with and without pseudo-rehearsal.   \textbf{Right}: Error bars are 95\% confidence intervals.
    }
    \label{fig:cmb_reward}
\end{figure}

\subsection{Controlling Agent}

As can be seen in Figure \ref{fig:cmb_reward}, continual learning in the controlling agent was observed in the task 1 results were inconclusive for task 2 or 3.  Task 1 (riverraid) in general shows a larger percent of total reward over training epochs when using pseudo-rehearsal as compared to without.  The plot of a particular replication illustrates that this is in large part due to a performance drop in later training for non-interleaved training.  For task 2 (tutankham), in general no reward was observed for this level of exposure, and so performance differences were unmeasureable.  Task 3, (crazy\_climber) however shows no significant difference between training conditions.  This null effect for continual learning could be due to a lack of complexity in this environment such that 1 million frames is sufficient to retrain the agent at every exposure, or that variance in reward over replications is too high to show any discernible differences in experiment conditions.

\section{Discussion and Conclusions}
Here we have shown that pseudo-rehearsal based methods are capable of supporting continual learning within the world model framework.  Our main focus was to illustrate that simulated rollouts generated from the internal model can be used to preserve its own temporal prediction across random task exposures with no explicit task labels.  The utility of this internal model, however, is generally measured in terms of its ability to inform a controlling agent.  Continual learning results when using the learned world model as input to modern policy gradient methods seem mixed.  

What has lead to these mixed results in the controlling network?  One possible explanation is that the world model architecture explored here and in \citet{ha2018} is not well suited for discreet action spaces.  \citet{henaff2017} suggest methods for modifying the loss function of the internal model for discreet actions, while \citet{kaiser2019} learn an action embedding. Similarly, \citet{ha2018} used evolutionary optimization on the controlling network as opposed to gradient descent.  Finally, in contrast to \citet{ha2018}, we did no explorations on temperature variation in the $M$ network for improving agent performance. In principle these methods could be adopted in the current work to improve performance.  

In this work we assumed a latent representation that is capable of encoding and decoding the features of new tasks.  There exists some preliminary work investigating continual learning in a VAE framework for classification \citep{rostami2019}.  Their approach essentially builds in a prior to the loss function that preserves the already existing latent space, which in principle would work in the current architecture.  Integrating this approach with the world model framework, however,  is left for future studies.  


Our main goal in this work was to explore the potential for continual learning within the world model architecture.  We have shown that pseudo-rehearsal is sufficient for continual learning within the internal model, or $M$ network, and that there exists methods for potentially improving the controlling agent's performance. Future work should focus on exploring a combined $V$ and $M$ framework that controls the latent space during continual learning, and allows for both discreet and continuous action embeddings.

\section*{Acknowledgments}
We thank James McClelland, Amarjot Singh, Charles Martin, and Mohammad Rostami for helpful feedback in the development and analysis of this work, and Jeffrey Krichmar, Emre Neftci, Risto Miikkulainen, Andrea Soltoggio, and Jean-Baptiste Mouret for conceptual discussions surrounding the work.  This material is based upon work supported by the United States Air Force \& DARPA under contract no. FA8750-18-C-0103. Any opinions, findings and conclusions or recommendations expressed in this material are those of the author(s) and do not necessarily reflect the views of the United States Air Force and DARPA.

\bibliographystyle{apalike}
\bibliography{cites}

\section{Supplemental Materials}

\subsection{Extra Figures}
\begin{figure}[h!]
    \includegraphics[width=\textwidth]{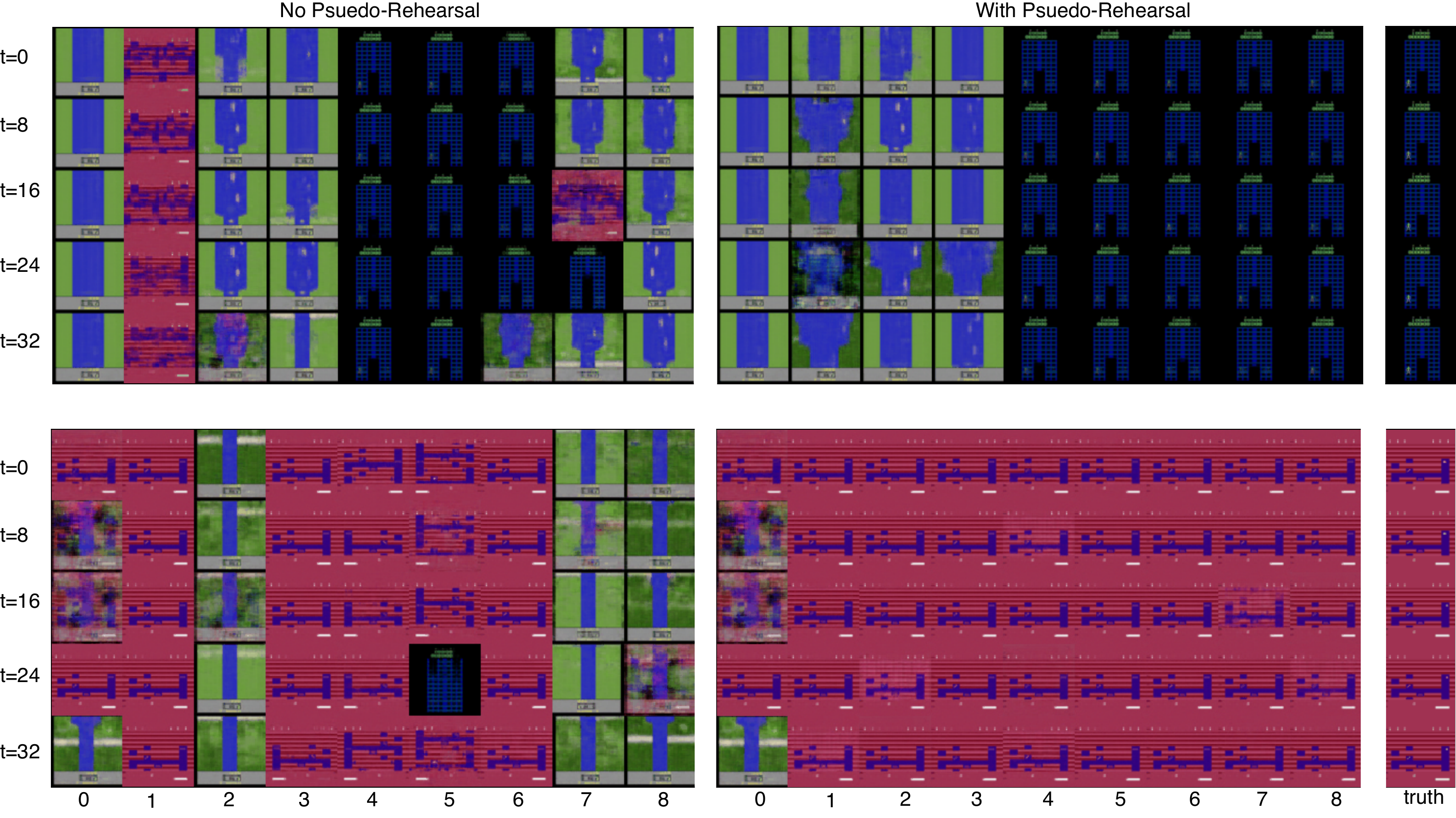}
    \caption{
    Reconstruction of test rollouts from tutankham (bottom) and crazy\_climber (top) using $M$ networks after training in each task exposure.  Left grid shows simulated rollouts when no pseudo-rehearsal was used in training, right grid shows simulated rollouts when trained with pseudo-rehearsal, and far right column shows real rollout from the environment.  Grid rows correspond to rollout images spaced 8 time steps (actual rollouts are spaced every 4 steps),  columns are rollouts generated after training is complete for each task exposure.
    }
    \label{fig:sup_cmb_recon}
\end{figure}

Here the reconstruction figures for the other 2 tasks from the main text are shown in Figure \ref{fig:sup_cmb_recon}.  As well as a schematic of the full network architecture used in Figure \ref{fig:full_arch}. 

Additionally included are animated gifs of simulated rollouts used for pseudo-rehearsal.  Their naming scheme `m.X.Y.gif` corresponds to the iteration X and the task that was trained during that iteration Y. 

\subsection{$M$ network optimization}
\label{sec:m_desc}
\begin{figure}[h!]
    \includegraphics[width=\textwidth]{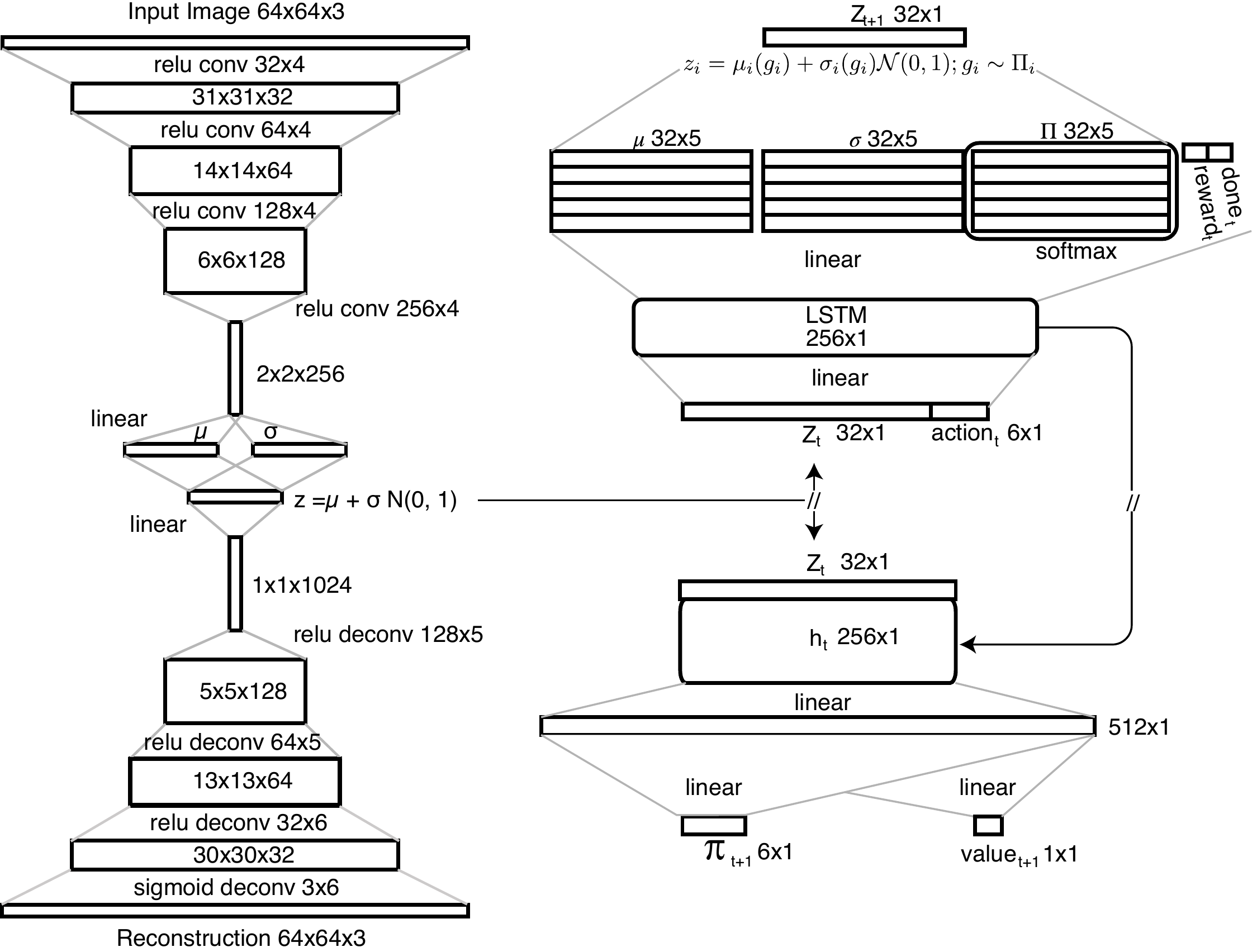}
    \caption{
    Schematic of full network.  Input starts on the top left through the $V$ network.  Once a particular $z$ vector is sampled, it is provided to the $M$ network (top right) along with the current action to produce distribution of potential next states.  The hidden state of the $M$ along with the current $z$ is provided as input to the controlling agent $C$ (bottom right).  This relatively simple network consists of a single linear layer of 512 units which splits into an actor and critic heads. Stops on the flow of gradients through the network are indicated by a `//' symbol.  Here `$\pi_t$' is an unnormalized vector of logits used to determine a distribution over potential actions, while `$action_t$' is a one-hot vector representing the particular action taken.
    }
     \small\textsuperscript{*} Figure adapted from \citet{ha2018}
    \label{fig:full_arch}
\end{figure}

A more detailed description of the $M$ optimization method is provided for extra clarity beyond the description in \citet{ha2018}. All symbols and indexes match those show in Figure \ref{fig:full_arch}.  Optimization of the LSTM and MDN is done comparing the output of the MDN to a $V$ encoded real observation ($z_{t+1}$) from 1 time step into the future predicated on the chosen action ($a_t$).  Specifically, the combined LSTM (with hidden state $h_t$) and MDN, referred to as the $M$ network, models each dimension (indexed by $i$) of the next latent state as the weighted sum of $G$ Gaussians: $P_i(z_{t+1} | x = (z_t, a_t, h_t)) = \sum^G_{g} \Pi_{g}(x) \mathcal{N}(\mu_{g}(x),\sigma_{g}(x))$.  Output units corresponding to each $\Pi_i$ set are normalized using a softmax, and each $\sigma_{g,i}$ unit is passed through an exponential function to ensure they are appropriate for the mixture model.  This set of Gaussian mixture models can be optimized using the average log likelihood over latent dimensions as the loss function: $\mathcal{L}_{GMM}(z_{t+1}, x) = 1/L\sum^L_{i} -\log(\sum^G_{g} \Pi_{g,i} \mathcal{N}(\mu_{g,i}(x), \sigma_{g,i}(x)))$.    Additionally, reward and terminal state units are optimized using mean squared error ($\mathcal{L}_{MSE}$) and binary cross entropy ($\mathcal{L}_{BCE}$), respectively.  Finally, the total loss optimized by the $M$ network is the average of the three losses $1/3 ( \mathcal{L}_{MSE}+ \mathcal{L}_{BCE}+\mathcal{L}_{GMM})$.

\newpage
\clearpage

%

\begin{figure}[b!]
    \includegraphics[width=\columnwidth]{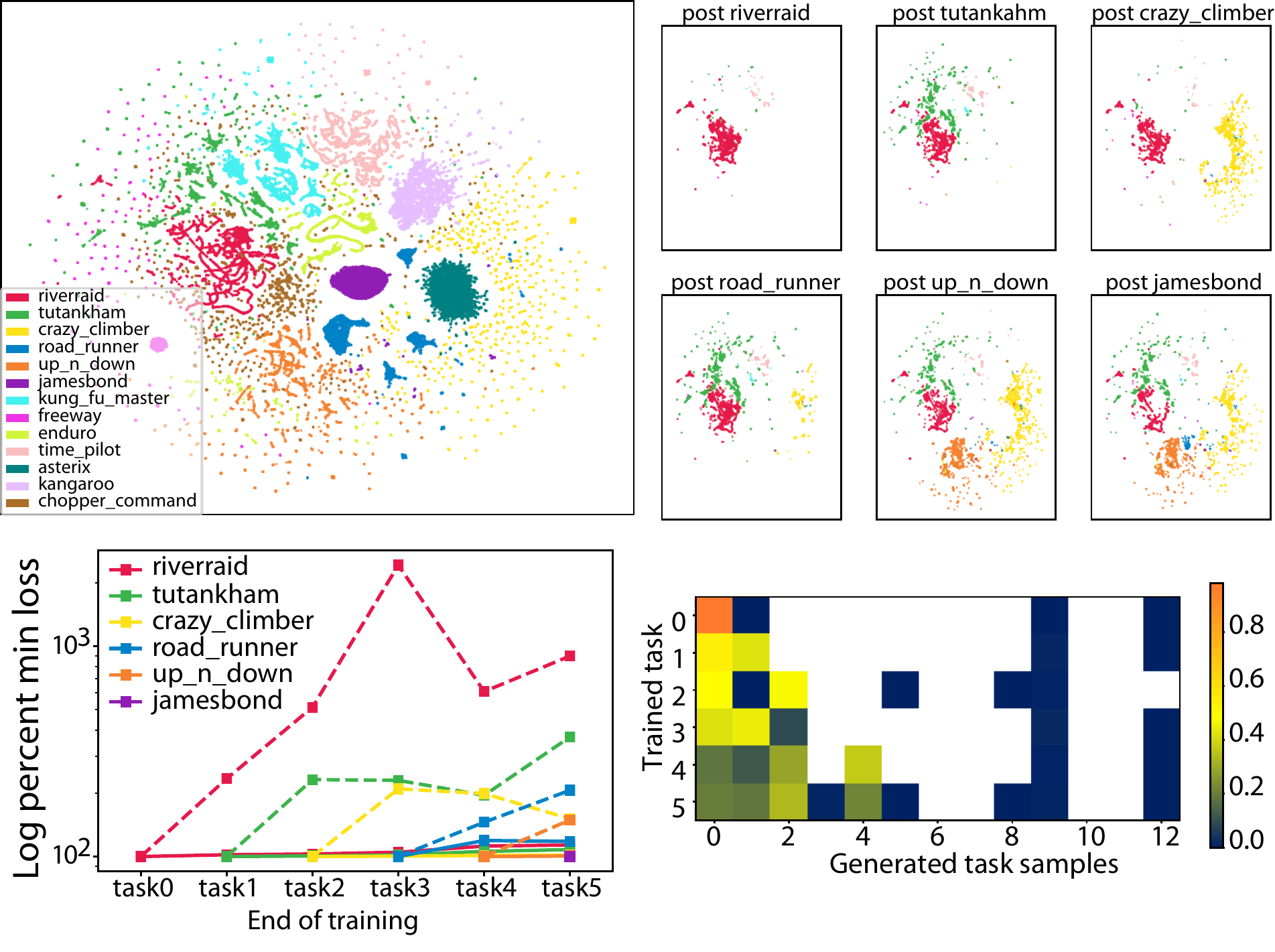}
    \caption{
    Results from supplemental experiment 1.  \textbf{Top-Left} plot shows 2D visualization of latent embedding using UMAP for all potential tasks.   \textbf{Top-Right} shows the pseudo-samples generated after training on a given task.  Samples are labeled using a KNN classifier based on the VAE's training data.  \textbf{Bottom-Left} plot  shows minimum loss after learning a given task.  Line colors show Atari game corresponding to the x-axis and order of learned tasks, e.g. `task0' corresponds to `riverraid'.  Solid lines show learning with pseudo-rehearsal and dashed line shows without.  Performance is percentage change from minimum loss within a given task, and y-axis is shown in the log scale. \textbf{Bottom-Right} plot shows the proportion of generated samples within each task for each iteration in the sequential training;  rows correspond iterations in the sequential training, and columns correspond to the proportion of samples labeled by the KNN classifier.  Each row sums to 1, and white squares correspond to cases which had a proportion of less than 0.004 of generated samples within a given iteration.  
    }
    \label{fig:emb_cmb}
\end{figure}

\subsection{Supplemental Experiments}
\label{sec:sup_exp}
Two supplemental experiments were done which focused on the $M$ and $V$ networks explicitly.  The first was done to investigate if the simulated rollouts can be attributed to task specific preservation. The second supplemental experiment investigated the larger potential for between task interactions and the ostensible capacity of the current architecture by training on a larger set of potential tasks.  

\subsubsection{Experiment 1}
 This experiment was done by first training the $V$ network to encode and decode a set of 13 Atari games.  This $V$ network was then used to encode samples from 6 (riverraid, tutankham, crazy\_climber, road\_runner, up\_n\_down, jamesbond) of those original games, and  sequentially trained the $M$ network on those 6 games.  During the course of training the $M$ network saw each game only once and was trained for 30 epochs or until test error did not decrease for 5 consecutive epochs.  No $C$ network was trained during this experiment and all rollouts were done using the random policy with a 0.5 probability of repeating the last randomly sampled action.

The simulated rollouts for each iteration of sequential training are shown in Figure \ref{fig:emb_cmb}.  Here the learned $V$ latent space is visualized in 2 dimensions using a UMAP embedding 
\citep{mcinnes2018} derived from a random set of test rollouts totaling 10,000 samples from each task.  Using this embedding a K-Nearest Neighbor (KNN) classifier  \citep{scikit-learn2011} is trained using the $V$ embedding of the same data, and then used to estimate a task label for pseudo-samples, which themselves are a random set totaling 10,000 samples taken from the rollouts interleaved during the experiment.  As can be seen in the bottom plot of Figure \ref{fig:emb_cmb} the pseudo-samples generated stay mostly within and distribute relatively well across trained tasks, even with no explicit task labels provided.  This sampling, however, clearly does deviate from uniform, and one consequence of obscuring task labels during training is that forcing a particular sampling over previous tasks becomes more difficult. 

Performance (similar to the main text, the loss function optimized during training) was assessed on a held out test set of rollouts for each task. This assessment was done on all previous tasks after training was complete for a given task, and was normalized by the minimum loss achieved for each respective task, i.e. the loss after initial training within a given task. In this way a measure of the degradation of performance as a function of learned tasks was achieved.  Bottom-left plot in Figure \ref{fig:emb_cmb} shows these metrics over training where, solid lines show performance when simulated rollouts were interleaved during training, and dashed lines show performance when no interleaving of simulated rollouts occurred. Clear catastrophic forgetting is seen in the non-interleaved case while relatively little reduction in loss is observed when simulated rollouts were interleaved.

In this controlled experiment of task exposures we found a significant relationship between percentage of pseudo-samples generated within a given task and the increase in loss within a iteration of the sequential learning experiment (Spearman rank correlation $r=-0.58, p=0.02$), suggesting that the more pseudo-samples used within a given task the smaller the increase in error.   One potential improvement from the current approach would be finding methods to enforce a smarter pseudo-sampling, however what the optimal sampling should be is an active area of research.

\newpage

\begin{figure}[b!]
    \includegraphics[width=\textwidth]{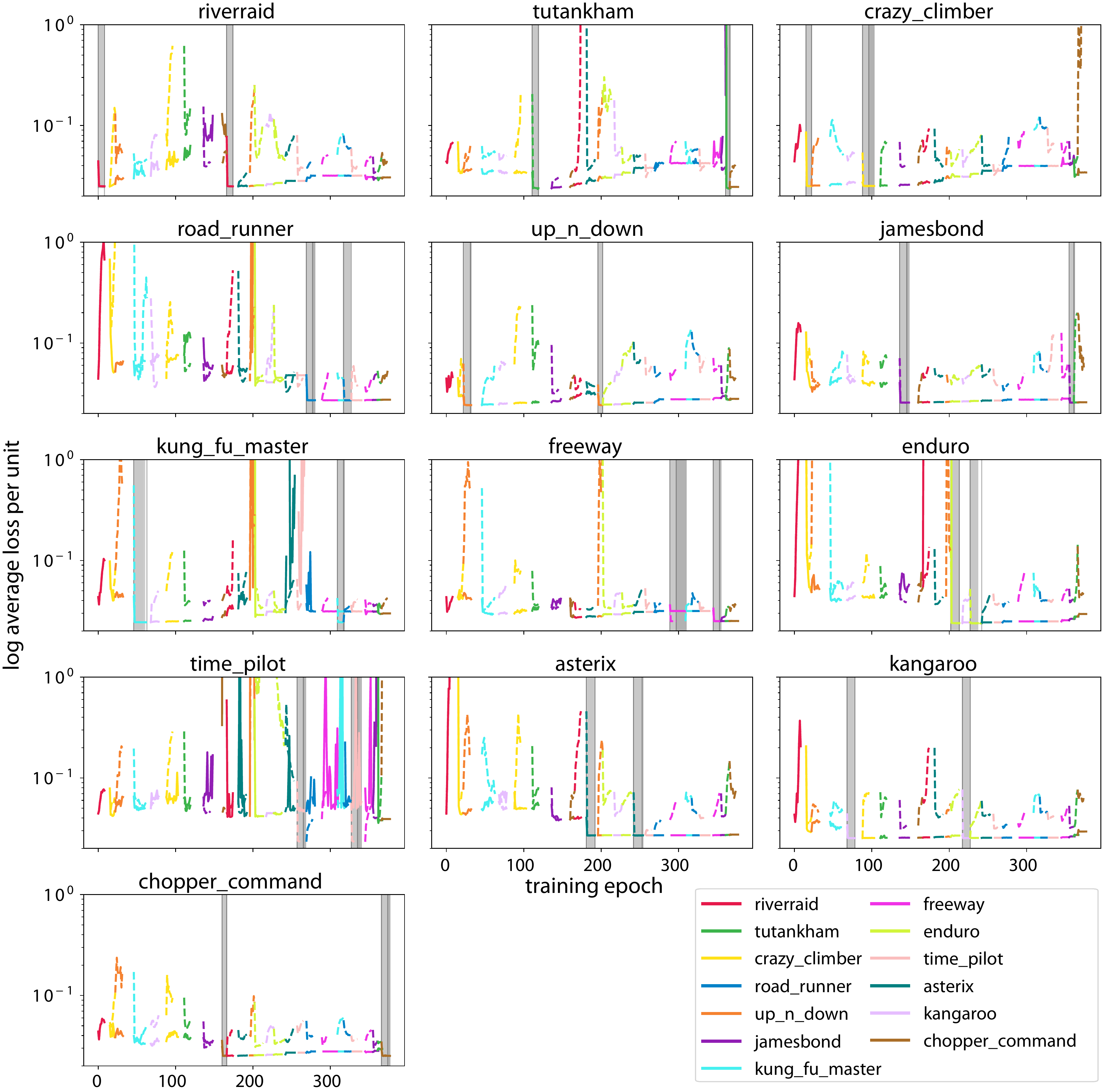}
    \caption{
    Results from supplemental experiment 2.  Here loss over training epochs for each task in random task exposure ordering is shown, similar to main text Figure \ref{fig:cmb_loss}.  Each sub-plot shows loss as measured in the hold-out set of the task label above it, lines colors correspond to the task that is currently being trained at a given epoch and grey overlays show when a that task is being exposed to its own training data. Solid lines show loss when using pseudo-rehearsal, dash lines without pseudo-rehearsal.     
    }
    \label{fig:noc_allenv}
\end{figure}

\begin{figure}[t!]
    \includegraphics[width=\textwidth]{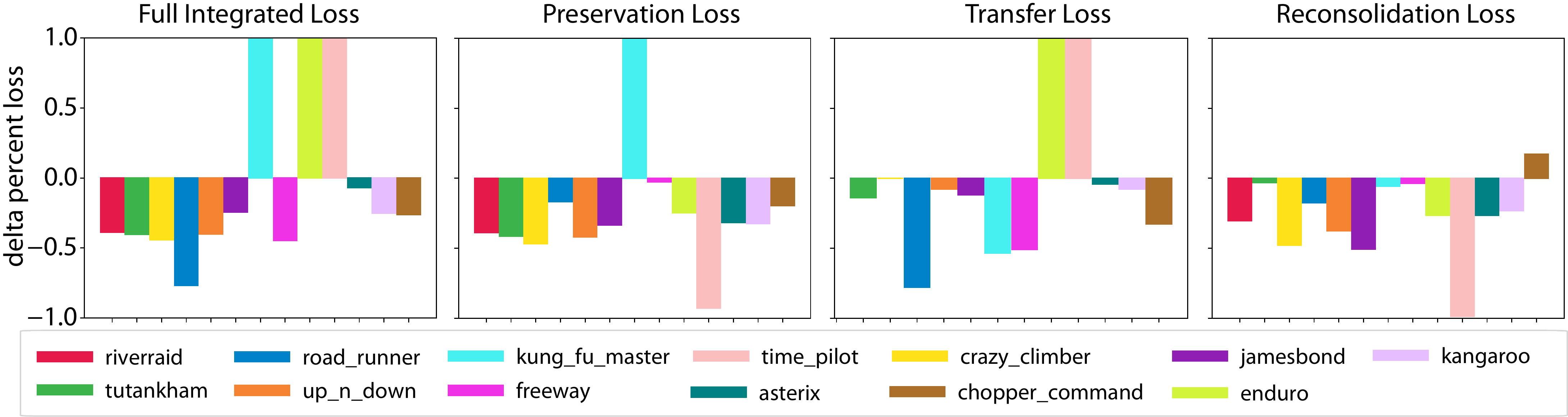}
    \caption{
\textbf{Left}: Pair-wise difference (with pseudo-rehearsal minus without)  in percent loss for each task when loss integration starts at first epoch of training.  \textbf{Middle-Left} plots the same as left except integration of loss starts on first exposure for each given task. \textbf{Middle-Right} plots same as left except loss integration done up to first exposure for a given task. \textbf{Right} plots same as left except loss integration starts from the last exposure for a given task.
    }
    \label{fig:noc_allenv_pctloss}
\end{figure}

\subsection{Experiment 2}

Here the $V$ network was again trained on the same 13 Atari games, and $M$ training was done in the mostly the same fashion as the experiment in the main text, i.e., a random order of task exposures, however, here training consisted of only two exposures with a sum total of 30 epochs per task.  Here we also used an early stopping criteria to try and limit over-training such that training was stopped in each task exposure if test error within a given task did not decrease over 5 epochs. No $C$ network was trained and rollouts were determined using the same random policy as done previously. 

Results shown in Figure \ref{fig:noc_allenv} illustrate the complex dynamics of the loss in the $M$ model when training over many potential tasks.  In general most tasks are learned as well as the 3 task experiment from the main text, however a larger accumulation of loss occurs for tasks that are not exposed until late in training. This can be seen in Figure \ref{fig:noc_allenv_pctloss} where the percent of integrated loss is plotted for each task.  The left plot in Figure \ref{fig:noc_allenv_pctloss} illustrates the pair-wise difference of with pseudo-rehearsal minus without pseudo-rehearsal loss as integrated from the first epoch of training in the firs task exposure.  Additionally, the middle-left plot labeled Preservation Loss shows the same metric when loss is integrated when a given task is first exposed, the middle-right plot labeled Transfer Loss illustrates the loss integrated up to a given task's first exposure, and the far right plot labeled Reconsolidation Loss shows the integrated loss starting from the second exposure until the end of training.  These metrics are similar to forward and backward transfer defined in \citet{lopez2017}.  Here we have a sign flip where Transfer Loss is analogous to the negative of Forward transfer (i.e., the influence that a given task exposure has on the performance of previous task exposures, in our case negative is better), and Preservation Loss is analogous to Backward transfer (i.e., the influence that a given task exposure has on future task exposures, again negative is better). 

These plots illustrate two main points.  The first is that the full integrated loss can effectively be broken down into Preservation Loss and Transfer Loss.  Here the potential for transfer between tasks can be seen in the middle-right plot of Figure \ref{fig:noc_allenv}. Some tasks' pseudo-rehearsal loss drop below the no-rehearsal loss before that task gets its first training exposure (e.g. negative values in the right most plot).  Similarly, it can be seen that several tasks' losses interact with each other in a complex way that would be hard to quantify but suggests the temporal prediction of those tasks are related, perhaps through the auto-encoder $V$.  

The second main result from this experiment is the illustration of continual learning in this large set of diverse tasks.  This not only occurs in the first task exposure, but seems to take root more completely after the second exposure, as measured by the Reconsolidation Loss (naming based on the memory literature \citep{alberini2013}).  One interpretation of this fact is that the more tasks the $M$ network is exposed to, the better it is at meta-learning.  This is generally in line with results from other meta-learning approaches which effectively exposure a network to batches of many related tasks to `initialize' the network's weights to be capable of quickly adapting to new incoming tasks \citep{nichol2018,finn2017}.  In contrast several tasks show Reconsolidation Loss comparable to when no pseudo-rehearsal is used.  Without running these simulations for longer, it is hard to determine if this is a measurement artifact due to lack of epochs for loss to accumulate after the final task exposure (e.g. tasks that had their last exposure near the end of training) or if this measure is manifesting some capacity in the current architecture.

\end{document}